\documentclass[sigconf]{acmart}

\usepackage{booktabs} 
\usepackage{array}
\usepackage{makecell}
\usepackage{amsmath}
\usepackage{amsfonts}
\usepackage{algorithm}
\usepackage{algorithmic}
\usepackage{multirow}
\usepackage{graphicx}
\usepackage{subfigure}
\usepackage{float}
\usepackage{bm}
\usepackage{hyperref}
\hypersetup{
    colorlinks=true,
    linkcolor=blue,
    filecolor=blue,      
    urlcolor=blue,
    citecolor=cyan,
}

\setcopyright{rightsretained}
\begin{document}
\begin{sloppypar}
\title{Introduction to The Dynamic Pickup and Delivery Problem Benchmark - ICAPS 2021 Competition}

\author{Jianye Hao*}
\affiliation
{\institution{Huawei Noah's Ark Lab}}

\author{Jiawen Lu*}
\affiliation
{\institution{Huawei Noah's Ark Lab}}

\author{Xijun Li*}
\affiliation
{\institution{Huawei Noah's Ark Lab}}

\author{Xialiang Tong*}\authornote{Authors are ranked in alphabetical order. Xialiang Tong is the corresponding author. His email is tongxialiang@huawei.com.}
\affiliation
{\institution{Huawei Noah's Ark Lab}}


\author{Xiang Xiang*}
\affiliation
{\institution{Huawei Noah's Ark Lab}}

\author{Mingxuan Yuan*}
\affiliation
{\institution{Huawei Noah's Ark Lab}}

\author{Hankz Hankui Zhuo*}
\affiliation
{\institution{Sun Yat-sen University}}


\begin{abstract}
The Dynamic Pickup and Delivery Problem (DPDP) is an essential problem within the logistics domain. So far, research on this problem has mainly focused on using artificial data which fails to reflect the complexity of real-world problems. In this draft, we would like to introduce a new benchmark from real business scenarios as well as a simulator supporting the dynamic evaluation. The benchmark and simulator have been published and successfully supported the ICAPS 2021 Dynamic Pickup and Delivery Problem competition participated by 152 teams. 
\end{abstract}

\maketitle

\section{Introduction}
Huawei, as one of the largest global communication device vendors, manufactures billions of products in hundreds of factories each year. A large amount of cargoes (including the raw materials, finished products and semi-finished products) need to be delivered among factories during the process of manufacturing. Due to the uncertainties of customers’ requirements and production processes, most delivery requirements cannot be fully decided in advance. The delivery orders, with the information including the pickup factories, delivery factories, the amount of cargoes and the time requirement, occur randomly and a fleet of homogeneous vehicles is periodically scheduled to serve these orders. Due to the large amount of deliver requests, even a small improvement of the logistics efficiency can bring significant benefits of finance. Therefore, it is of great significance to develop an efficient optimization algorithm to dispatch orders and plan the route of vehicles.

Masking historical datasets from real system is published with the simulator for algorithm development. The dataset contains 30 days of historical data, with 2000\textasciitilde5000 orders and 100\textasciitilde200 vehicles per day. An order can be served by several vehicles when its loading demand exceed the capacity of one vehicle.
A simulator with result checking method is also provided to evaluate the performance of the algorithm over the planning horizon (e.g., one day). In the simulator, a day is split into several time intervals with the same duration (e.g., $T=144$ intervals and time interval $\Delta t = 10\; min$). At the beginning of each time interval, the simulator will pass the latest environment information to the developed algorithm. After that, the scheduling results will be sent back to the simulator from the algorithm, to allow the simulator simulating the delivery process. The process will continuously be carried out until all orders are completed.
To the best of our knowledge, this is the first set of dataset and simulator derived from real industrial scenes for the dynamic pickup and delivery problem, which enables related research more practical and efficient.

As mentioned above, to promote the research and application on real world problems, we organized the Dynamic Pickup and Delivery Problem competition at the 2021 International Conference on Automated Planning and Scheduling (ICAPS 2021 DPDP). In this competition, participants were expected to design algorithms from any perspectives (e.g., combinatorial optimization, heuristics methods, etc.) to schedule vehicles to serve orders generated dynamically which contain the demand to deliver cargoes from pickup factories to destinations. The overall objective is to minimize the overtime of orders as well as the average traveling distance of vehicles.

In Section 2, we will give detailed information for the benchmark, which contains detailed description for DPDP, the formulation and corresponding evaluation criteria of the problem, and the utilization methods of the benchmark.
In Section 3, we will introduce the related work on this problem.
In Section 4, a summary of ICAPS 2021 DPDP Competition is discussed.

\section{Benchmark Description}
\subsection{Introduction to DPDP}
Like information aforementioned, a large amount of cargoes need to be delivered among factories during the manufacturing, and each factory could be either pickup point or delivery point. Due to the uncertainties of customers’ requirements and production processes, most delivery requirements cannot be fully decided beforehand. The delivery orders, with the information including the pickup factories, delivery factories, the amount of cargoes and the time requirement, occur randomly and a fleet of homogeneous vehicles is periodically scheduled to serve these orders.

\begin{figure}[h!]
    \centering
    \includegraphics[width=8cm]{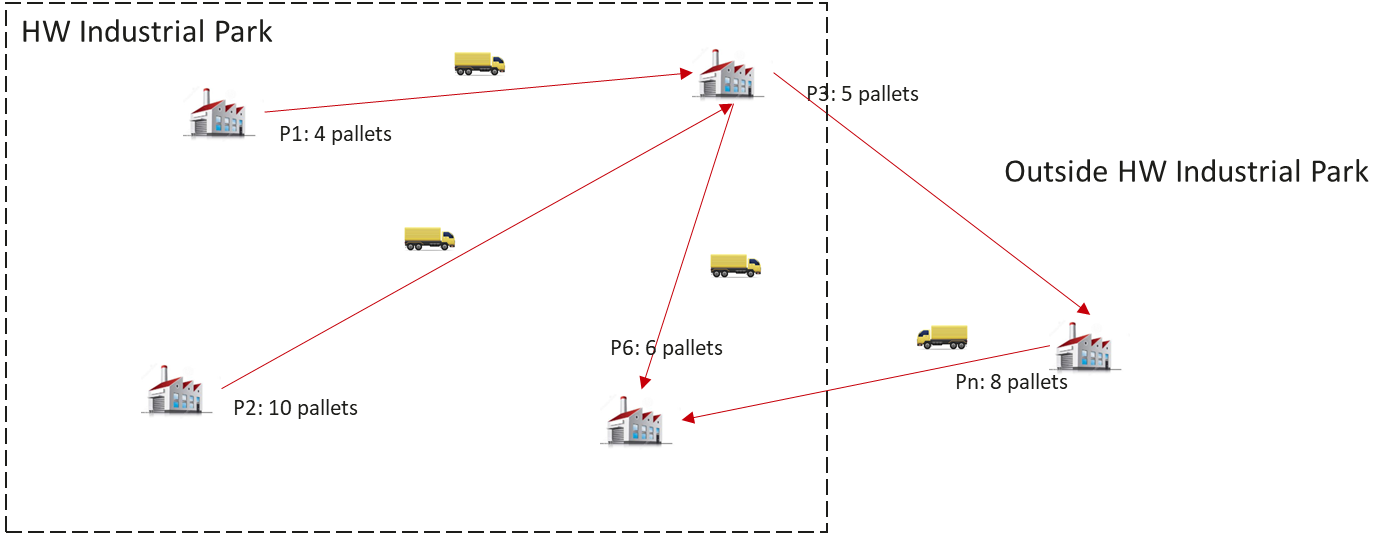}
    \caption{An Example of DPDP}
    \label{fig:DPDP_Example}
\end{figure}

Like shown in \textbf{Figure} \ref{fig:DPDP_Example}, orders generate at factories, which contain cargoes to be sent from one factory to another. The algorithm developed should decide which vehicle is assigned to carry out this task, and also which route the vehicle should select.

\subsection{DPDP Formulation and Evaluation}
Generally, DPDP is formulated in several parts: input, output, constraints, and objective. Details are given below.
\\
\textbf{Input:}
    \begin{itemize}
        \item A road network $G = (F,A)$, which is a complete directed graph. $F$ is the set which consists of multiple nodes (standing for factories in this competition): $\{F_i|i=1,\cdots,M\}$, and $A=\{(i,j)|i,j\in M\}$ is the arc set. Each arc $(i, j)$ is associated with a non-negative transportation distance $d_{ij}$ and transportation time $t_{ij}$ from node $i$ to node $j$.

        \item  An order set $O=\{o_i|i=1,\cdots,N\}$, in which each order $o_i=(F_p^i,F_d^i,q^i,t_e^i,t_l^i)$, where $F_p^i$ and $F_d^i$ represent the pickup and delivery node respectively, $q^i$ is the amount of cargoes to be delivered, which is expressed as: $q^i = (q_{standard}^i, q_{small}^i, q_{box}^i)$, where $q_{standard}^i$ stands for the standard pallet amount, $q_{small}^i$ stands for the small pallet amount, and $q_{box}^i$ stands for the box amount, $t_e^i$ refers to the creation time and $t_l^i$ is the committed completion time (Note that the committed completion time refers to the competition time of each order, including the vehicle travel time, dock approaching time, dock queuing time and loading/unloading time). \textbf{$1$ standard pallet equals $2$ small pallets, and $1$ small pallet equals $2$ boxes.} These orders come randomly.
        
        \item A fleet of homogeneous vehicles $V=\{v_k|k=1,\cdots,K\}$. Each vehicle $v_k$ is associated with a loading capacity and work shift of drivers. The initial positions of vehicles are randomly assigned to nodes.
        
        \item $M$ nodes (factories) $\{F_i|i=1,\cdots,M\}$ . Each node has a limited number of cargo docks for loading and unloading, as well as work shift constraint. If all docks are in use, the newly arriving vehicles must wait for their release. And also the loading and the unloading operations can only be performed during work shift.
        
        \item Dock approaching time, the time for a vehicle between arrival at the factory and the allocation the dock without considering queuing, e.g., $30$ minutes.
        
        \item Loading and unloading time, e.g., if there are $q$ standard pallets in the vehicle, then the loading time $t_p=\omega \times q$ and unloading time $t_d=\omega \times q$,\; $\omega = 180 \; seconds/standard\ pallet$.
    \end{itemize}

\textbf{Output:}
    \begin{itemize}
        \item The order assignment plan and the route of each vehicle, e.g., for vehicle $v_k$, its route plan: $\prod_k = \left\{ n_1^k,\ n_2^k,\cdots,\ n_{l_k}^k \right\}$, where $n_i^k$ is the $i$th factory the vehicle should visit.
    \end{itemize}

\textbf{Constraints:}
    \begin{itemize}
        \item Order fulfillment. All orders need to be served.
        \item Committed completion time. Orders need to be completed within the committed completion time, e.g., $4$ hours. Otherwise, the penalty cost would arise.
        \item Order split. Splitting is only allowed for orders that exceed the vehicle's loading capacity. The smallest unit of the order is not allowed to be divided. For example, if an order contains $13$ standard pallets, $7$ small pallets and $1$ box, the smallest units are $1$ standard pallet or $1$ small pallet or $1$ box.
        \item Loading capacity of the vehicle. For each vehicle $v_k$ , the total capacity of loaded cargoes cannot exceed the maximum loading capacity Q, e.g., $12$ standard pallets per vehicle.
        \item Work shift of drivers, e.g., $8:30-12:00,\ 13:30-18:00$.
        \item Last In First Out (LIFO) loading constraint. e.g., if orders $o_1$ and $o_2$ are assigned to the same vehicle $v_k$, where the pickup and delivery node of $o_1$ are $F_p^{1}$ and $F_d^{1}$, the pickup and delivery node of $o_2$ are $F_p^{2}$ and $F_d^{2}$, then the route plan $\left\{ F_p^{1},\ F_p^{2},\ F_d^{1},\ F_d^{2} \right\}$ violates the LIFO constraint, while $\left\{ F_p^{1},\ F_p^{2},\ F_d^{2},\ F_d^{1} \right\}$ does not.
        \item Limited number of cargo docks of each node: e.g., if node $F_1$ contains $3$ cargo docks and $4$ vehicles arrive, then the latest vehicle has to wait until one dock turns available.
        \item Loading and unloading is subject to the first come, first serve rule. If multiple vehicles arrive at the same node at the same time while the node has only one vacant cargo dock, then the node will randomly select one vehicle for loading and unloading. Note that the above situation will not happen in the reality, but may occur in the simulated environment.
    \end{itemize}
    
\textbf{Objectives:}
\\ This problem includes two objectives to optimize.
    \begin{itemize}
        \item The first objective is to minimize the total timeout of orders denoted as $f_1$:
        $$f_1 = \sum_{i=1}^{N} \max(0,\ a_d^i - t_l^i)$$
        Where $a_d^i$ and $t_l^i$ denote the arrival time to the destination and the committed completion time of order $o_i$ respectively, and $N$ is the total amount of orders.
        
        \item The second objective is to minimize the average traveling distance of vehicles denoted as $f_2$:
        $$f_2 = \frac{1}{K} \sum_{k=1}^{K} \sum_{i=1}^{l_k-1} d_{n_i^k,n_{i+1}^k}$$
        Considering that the route plan of vehicle $v_k$ is $\prod_k = \left\{ n_1^k,\ n_2^k,\cdots,\ n_{l_k}^k \right\}$, where $l_k$ is the total amount of factories vehicle $v_k$ should visit, and $d_{n_i^k,n_{i+1}^k}$ is the distance from node $n_i^k$ to node $n_{i+1}^k$
        
        \item The overall objective is acquired via:
        $$\min f = \lambda \times f_1 + f_2$$
        Where $\lambda$ is a pretty large positive constant, i.e., we focus much more on the optimization of $f_1$
    \end{itemize}

\subsection{Utilization of the Benchmark}
We'd like to introduce the utilization of our simulator and dataset in this part. For downloading and supplementary information for them, please access via this link: \href{https://github.com/huawei-noah/xingtian/tree/master/simulator/dpdp_competition}{DPDP Simulator and Dataset}. \\
\textbf{Description of the Dataset:}
\\
The dataset is divided into several data types: orders, vehicles, route map, and factory information.
\begin{itemize}
    \item Orders \\
    Each order file contains one day's order information, with a file name in format \textit{"order\_count} + \textit{file\_number"}. \textit{order\_count} denotes the order quantity, and \textit{file\_number} is used to differentiate the order data for different days. For example, file \textit{50\_1} means that this is the first file with $50$ orders per day. The \textit{order\_count} is ranging from $50$ to $4000$ which helps the users test their algorithm in different order quantities. The attributes in order file are explained in \textbf{Table} \ref{tab:description_order}.
    \begin{table}[h]
        \centering
        \caption{Description of Order Data}
        \label{tab:description_order}
        \begin{tabular}{|c|c|c|}
            \hline
            \textbf{Column} & \textbf{Description} & \textbf{Example} \\
            \hline
            order\_id & ID of order & 0003480001 \\
            \hline
            q\_standard & Standard pallet amount & 1 \\
            \hline
            q\_small & Small pallet amount & 2 \\
            \hline
            q\_box & Box amount & 1 \\
            \hline
            demand & \makecell{Total standard pallet\\amount, calculated as:\\ $q\_standard + 0.5 \times$ \\ $q\_small + 0.25 \times q\_box$} & 1.75 \\
            \hline
            creation\_time & \makecell{Order creation time\\ (\%H:\%M:\%S)} & 00:03:48 \\
            \hline
            \makecell{committed\_\\completion\_time} & \makecell{Order committed\\ completion time\\ (\%H:\%M:\%S)} & 04:03:48 \\
            \hline
            load\_time & \makecell{Order loading time\\ (unit: second)} & 120 \\
            \hline
            unload\_time & \makecell{Order unloading time\\ (unit: second)} & 120 \\
            \hline
            pickup\_id & ID of pickup factory & \makecell{2445d4bd004c457d\\95957d6ecf77f759} \\
            \hline
            delivery\_id & ID of delivery factory & \makecell{b6dd694ae05541db\\a369a2a759d2c2b9} \\
            \hline
        \end{tabular}
    \end{table}
    \\
    \item Vehicles \\
    Each file contains available vehicles information, with a file name in format \textit{"vehicle\_info} + \textit{vehicle\_count"}. \textit{vehicle\_count} denotes the number of vehicles available which is ranging from $5$ to $100$, make it convenient for users to test their algorithm in different vehicle quantities. The attributes are explained in \textbf{Table} \ref{tab:description_vehicle}.
    
    \begin{table}[h]
        \centering
        \caption{Description of Vehicle Data}
        \label{tab:description_vehicle}
        \begin{tabular}{|c|c|c|}
            \hline
            \textbf{Column} & \textbf{Description} & \textbf{Example} \\
            \hline
            car\_num & ID of vehicle & V\_1 \\
            \hline
            capacity & \makecell{Capacity of vehicle\\ (unit: standard pallet)} & 15 \\
            \hline
            operation\_time & \makecell{Operation time\\ of vehicle\\ (unit: hour)} & 24 \\
            \hline
            gps\_id & \makecell{ID of GPS\\ equipment} & G\_1 \\
            \hline
        \end{tabular}
    \end{table}
    
    \item Route Map \\
    The route\_map file offers the distance and travel time between each pair of factories. Details are given in \textbf{Table} \ref{tab:description_route_map}.
    \begin{table}[]
        \centering
        \caption{Description of Route Map}
        \label{tab:description_route_map}
        \begin{tabular}{|c|c|c|}
            \hline
            \textbf{Column} & \textbf{Description} & \textbf{Example} \\
            \hline
            route\_code & ID of route & \makecell{e7eeb0e4-a7c7\\-11eb-8344-\\84a93e824626} \\
            \hline
            start\_factory\_id & \makecell{Departure factory\\ID of the route} & \makecell{7782ed919d8f4dd6\\a1fb220dacd73445} \\
            \hline
            end\_factory\_id & \makecell{Terminal factory\\ ID of the route} & \makecell{43a4215be06543c1\\985c1e9460dec52d} \\
            \hline
            distance & \makecell{Distance of the route\\(unit: km)} & 76.0 \\
            \hline
            time & \makecell{Transportation time\\of the route\\(unit: second)} & 10140 \\
            \hline
        \end{tabular}
    \end{table}
    
    \item Factory Information \\
    The factory\_info file gives the location and port (dock) information. Details are given in \textbf{Table} \ref{tab:factory_info}.
    \begin{table}[]
        \centering
        \caption{Description of Factory Data}
        \label{tab:factory_info}
        \begin{tabular}{|c|c|c|}
            \hline
            \textbf{Column} & \textbf{Description} & \textbf{Example} \\
            \hline
            factory\_id & ID of factory & \makecell{9829a9e1f6874f28\\b33b57a7a42bb49f} \\
            \hline
            longitude & Longitude of factory & 116.6259 \\
            \hline
            latitude & Latitude of factory & 40.2204 \\
            \hline
            port\_num & \makecell{Number of ports used\\for loading and unloading\\of vehicle cargoes} & 6 \\
            \hline
        \end{tabular}
    \end{table}
\end{itemize}

\begin{figure}[htp]
    \centering
    \includegraphics[width=8cm]{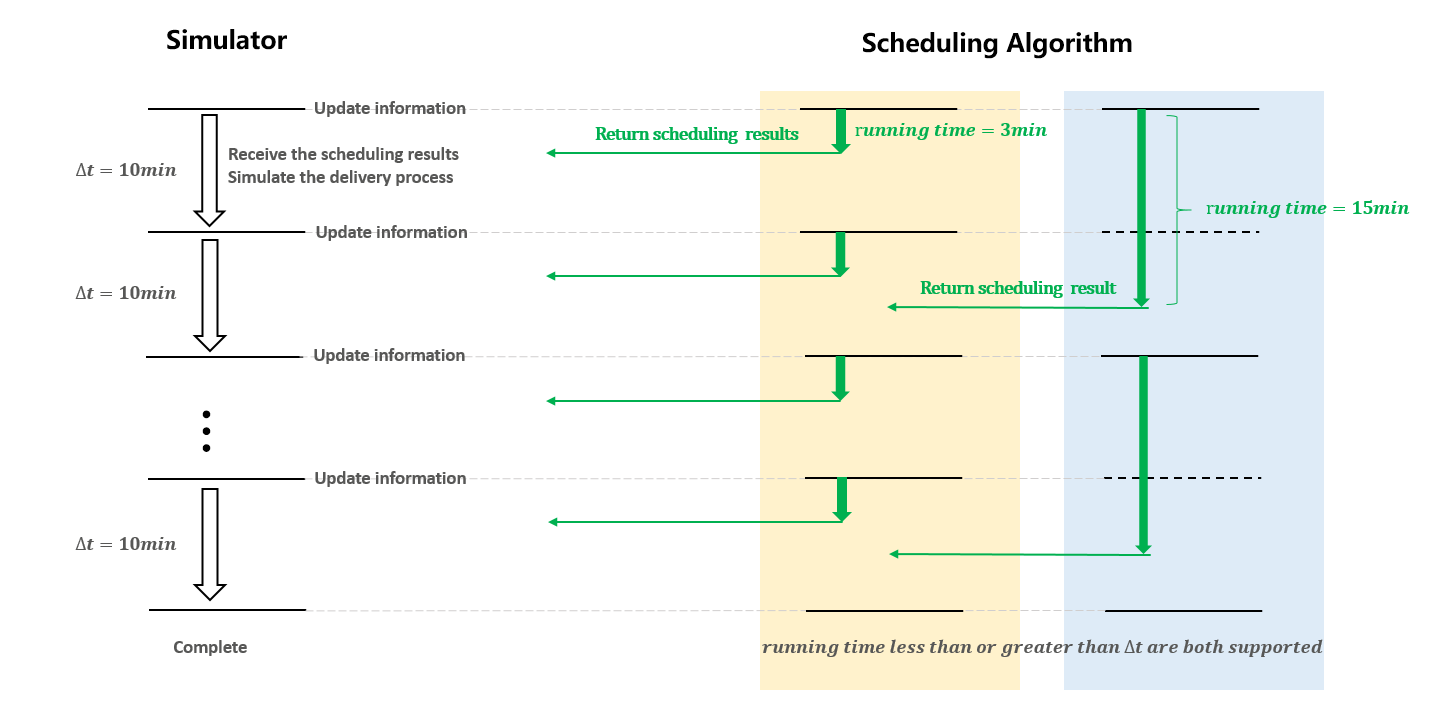}
    \caption{Interaction between the Simulator and Algorithm}
    \label{fig:interaction_simulator_algorithm}
\end{figure}

\begin{figure}[htp]
    \centering
    \includegraphics[width=8cm]{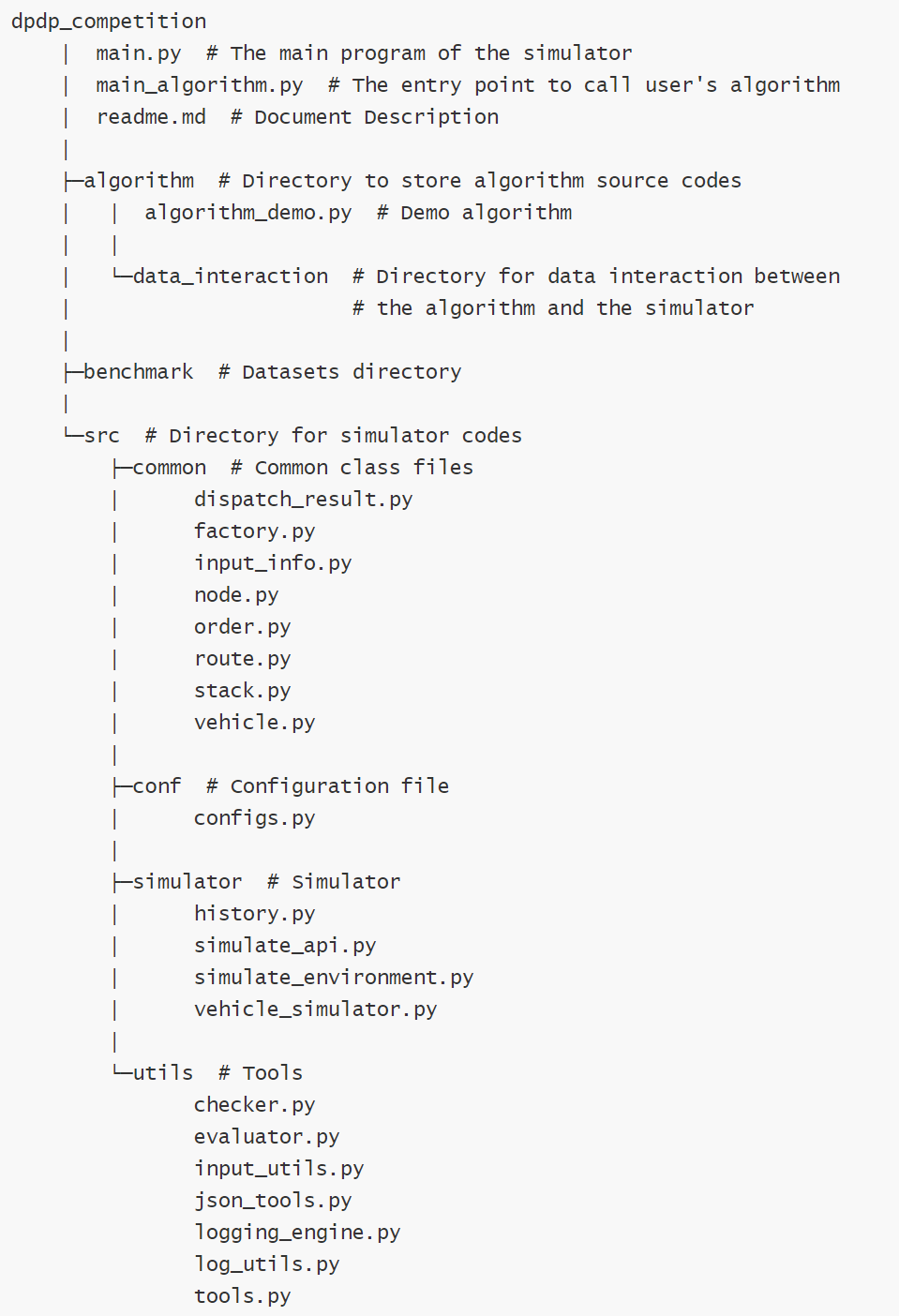}
    \caption{Directory Structure of the Simulator}
    \label{fig:directory_structure_simulator}
\end{figure}

\noindent \textbf{Description of the simulator:}
\\
The simulator with result checking function is provided to evaluate the performance of the algorithm over the planning horizon, e.g., one day. In the simulator, a day is split into $T$ time intervals of the same duration, e.g., $T=144$ intervals and time interval $\Delta t = 10\; min$. A flowchart is given in \textbf{Figure} \ref{fig:interaction_simulator_algorithm} for reference and the directory structure of the simulator is displayed in \textbf{Figure} \ref{fig:directory_structure_simulator}.

Below is detailed description for utilization of the simulator from simulator perspective.

\begin{itemize}
    \item \textbf{Interaction between Algorithm and Simulator} \\
    To start with, the simulator reads the selected test instances which can be modified in \texttt{Configs.py}. Then the simulator performs simulation at a fixed interval of 10 minutes until all orders in the instance are completed. \\
    In each round of simulation, the simulator outputs the vehicle and the order data required by the algorithm to the \texttt{algorithm/data\_interaction} folder in JSON format. Next, it calls the main program of the algorithm prefixed with "main\_algorithm", e.g., \texttt{main\_algorithm.py}, \texttt{main\_algorithm.java}, etc. When the algorithm runs, it starts to read JSON files, dispatch orders and output the dispatch result to the algorithm/data\_interaction folder in JSON format and prints string "SUCCESS" to the console as the indicator for the simulator to determine whether the algorithm is successfully executed. If the successful indicator is obtained, the simulator would read the output JSON files and do the verification. After passing the verification, it continues to simulate next round. \\
    The running time of the algorithm per round is limited to 10 minutes. The simulator will exit once the algorithm times out.
\end{itemize}

And below describes the interaction methods from algorithm perspective.
\begin{itemize}
    \item \textbf{Read the Input JSON Files} \\
    The simulator outputs the latest vehicle information in \texttt{vehicle\_info.json}, order items to be allocated in \texttt{unallocated\_order\_items.json} and ongoing order items in \texttt{ongoing\_order\_items.json}. Apparently, \texttt{unallocated\_order\_items.json} and \texttt{ongoing\_order\_items.json} have the same format. The details of these files are in \textbf{Table} \ref{tab:vehicle_json} and \textbf{Table} \ref{tab:order_json}.
    Some key concepts are explained here:
    \begin{itemize}
        \item Item, which refers to the smallest indivisible unit in an order. For example, if an order contains 2 standard pallets and 1 small pallets, there are 3 individual items totally: item 1 containing one standard pallet, item 2 containing one standard pallet and item 3 containing 1 small pallet.
        \item Item status 0 means initialization.
        \item Item status 1 means the item is generated.
        \item Item status 2 means the item has been loaded.
        \item Item status 3 means the item is delivered.
        \item Order status, which is the minimum value of all corresponding items statuses. For example, an order includes two items, namely, item 1 containing 1 small pallet and item 2 containing 1 box. The status of item 1 is 3 and item 2 is 2, so the overall order status is 2.
        \\
    \end{itemize}
    
    \begin{table}[]
        \centering
        \caption{Details of \texttt{vehicle\_info.json}}
        \label{tab:vehicle_json}
        \begin{tabular}{|c|c|c|}
            \hline
            \textbf{Column} & \textbf{Description} & \textbf{Type} \\
            \hline
            id & ID of vehicle & str \\
            \hline
            operation\_time & \makecell{Operation time of vehicle\\(unit: hour)} & int \\
            \hline
            capacity & \makecell{Capacity of vehicle\\(unit: standard pallet)} & int \\
            \hline
            update\_time & \makecell{Update time of the\\current position and\\status of the vehicle\\(unit: Unix timestamp)} & int \\
            \hline
            cur\_factory\_id & \makecell{Factory ID where the\\ vehicle is currently located.\\ Value is "" if the vehicle is \\currently not in any factory.} & str \\
            \hline
            \makecell{arrive\_time\_at\_\\current\_factory} & \makecell{Time when the vehicle\\ arrives at the current factory\\(unit: Unix timestamp)} & int \\
            \hline
            \makecell{leave\_time\_at\_\\current\_factory} & \makecell{Time when the vehicle\\ leaves the current factory\\(unit: Unix timestamp)} & int \\
            \hline
            carrying\_items & \makecell{List of items loaded\\ on the vehicle in\\ the order of loading} & [str, str,$\cdots$] \\
            \hline
            destination & \makecell{Current destination of the\\ vehicle. Once determined, the\\ destination cannot be changed\\ until the vehicle has arrived.\\ The destination is None when\\ the vehicle is parked.} & dict \\
            \hline
        \end{tabular}
    \end{table}
    
    \begin{table}[]
        \centering
        \caption{Details of Order Items JSON File}
        \label{tab:order_json}
        \begin{tabular}{|c|c|c|}
            \hline
            \textbf{Column} & \textbf{Description} & \textbf{Type} \\
            \hline
            id & ID of item & str \\
            \hline
            type & Pallet type & str \\
            \hline
            order\_id & ID of order & str \\
            \hline
            demand & \makecell{Total amount of the order\\ in standard pallet unit} & double \\
            \hline
            pickup\_factory\_id & ID of pickup factory & str \\
            \hline
            delivery\_factory\_id & ID of delivery factory & str \\
            \hline
            creation\_time & \makecell{Creation time of the\\ corresponding order\\ (unit: Unix timestamp)} & int \\
            \hline
            \makecell{committed\_\\completion\_time} & \makecell{Committed completion time\\ of the corresponding order\\ (unit: Unix timestamp)} & int \\
            \hline
            load\_time & \makecell{Loading time of item\\(unit: second)} & int \\
            \hline
            unload\_time & \makecell{Unloading time of item\\(unit: second)} & int \\
            \hline
            delivery\_state & Item status 0\textasciitilde3 & int \\
            \hline
        \end{tabular}
    \end{table}
    
    \item \textbf{Dispatch Orders}
    \begin{itemize}
        \item When the vehicle $v$ arrives at the factory $f$, the pickup and delivery list of $v$ in $f$ will be generated immediately. The vehicle can only be loaded and unloaded according to the list. We can only change the pickup and delivery items of $v$ in $f$ when the corresponding pickup and delivery list is not generated.
        \item Algorithm can reallocate the item to different vehicles as long as this item is not displayed on the  pickup and delivery list.
        \item Considering that the input is the items, the algorithm needs to pay attention to the order splitting constraint. If the order does not exceed the vehicle's capacity (vehicles are homogeneous), the order cannot be split.
        \item The algorithm can control the order release. For example, if order A is generated at $t_1$, the algorithm can delay the allocation until $t_2 <= t_1 + 4h$. Note: If an order has been generated for more than 4 hours but is still not dispatched, the simulator will exit.
        \item The traveling distance and time required for the algorithm can be obtained only from the distance and time matrix between factories in the benchmark. Do not calculate the distance and time based on the longitude and latitude. If the vehicle $v$ is in transit,  $v$ must have a destination factory $f$, and the simulator will give the estimated time for $v$ to arrive at $f$. The algorithm can plan the route of vehicle $v$ based on the destination $f$.
        \item Assume that the simulator sends the latest status information of all vehicles to the algorithm at $t_1$, and the algorithm returns the dispatching result at $t_2$. If the algorithm runs for a long time, the status of vehicles will change greatly after passing  the time interval $[t_1, t_2]$. As a result, the dispatching result is inconsistent with the actual situation, which affects the running of vehicles. Currently, the running time of the algorithm is limited to 10 minutes.
        \\
    \end{itemize}
    
    \item \textbf{Output Required JSON Files} \\
    The algorithm needs to output two JSON files: \texttt{output\_destination.json} and \texttt{output\_route.json}. These two files have similar format, which could be explored through the readme file of the simulator. The attributes of them are explained in \textbf{Table} \ref{tab:attributes_output}.
    \begin{table}[]
        \centering
        \caption{Attribute of Output files}
        \label{tab:attributes_output}
        \begin{tabular}{|c|c|c|}
            \hline
            \textbf{Column} & \textbf{Description} & \textbf{Type} \\
            \hline
            factory\_id & ID of factory & str \\
            \hline
            lng & Longitude of factory & double \\
            \hline
            lat & Latitude of factory & double \\
            \hline
            delivery\_item\_list & \makecell{List of items unloaded\\ from the vehicle} & [str, str, $\cdots$] \\
            \hline
            pickup\_item\_list & \makecell{List of items loaded\\ on the vehicle} & [str, str, $\cdots$] \\
            \hline
            arrive\_time & \makecell{Time to reach\\ the factory} & int \\
            \hline
            leave\_time & \makecell{Time to leave\\ the factory} & int \\
            \hline
        \end{tabular}
    \end{table}
\end{itemize}

\section{Related Works}
In academic aspect, DPDP is a complex variant of Vehicle Routing Problem (VRP), which has been proved to be a NP-Hard combinatorial optimization problem \cite{toth2014vehicle}. Compared to VRP, DPDP is more complex considering there are various constraints like vehicle capacity constrain, Last-In-First-Out constraint, time window constrain, split demand constraint, etc. On top of this, the orders come in a random manner which is hard to predict and acquire beforehand. There are two popular methods to solve DPDP. The first one is exact methods, which consumes too much time and doesn't meet the need for problems at large scales \cite{savelsbergh1998drive}. The other one is heuristic methods, which divide the problem into a series of static problems and conquer them individually \cite{gendreau2006neighborhood, mitrovic2004double}. However, a series of sub-optimal solutions don't mean an optimal solution globally. Some methods predict the distribution of future orders to improve the solution in sub-problem \cite{thomas2007waiting,ghiani2009anticipatory}.
With the development of deep learning and reinforcement learning, some researchers design learning-based methods to solve DPDP. \cite{ma2021a} proposes a novel novel bi-level reinforcement learning agents optimization framework. The upper-level agents determines whether to release orders of the given static problem to the lower-level agent and the lower-level agent is responsible for order dis-patching and vehicle routing planning by improving an initial solution iteratively. \cite{li2021learning} designs and end-to-end reinforcement learning solution using double deep graph networks incorporating attention-based graph embedding at industrial scale.

\section{Brief Summary of ICAPS 2021 Dynamic Pickup and Delivery Problem Competition}
This competition was organized by Huawei Noah's Ark Lab with Sun Yat-sen University at 2021 International Conference on Automated Planning and Scheduling. To the best of our knowledge, this competition is the first one to offer comprehensive support, from simulator, dataset to execution platform, in this field.
\\
Totally 843 participants world-widely joined the competition and 152 teams were formed. In general, our participants used miscellaneous methods like heuristic methods, combinatorial optimization, reinforcement learning, and so on and so forth. Actually these methods are common ones often explored by both academic and industrial fields. Though in many cases heuristic methods are relatively easy to find out good acceptable solutions in a highly efficient manner, we are delighted to see there are still many participants trying in other perspectives like reinforcement learning, and we hope this could be a promising start to encourage more people to discover different aspects and core of the problem via sorts of methods. 
\\
Finally, 3 teams outperformed all other teams to win the prizes, whose algorithms utilize heuristic methods mainly. One of the awarded teams utilized Variable Neighbourhood Search method for finding out the route plans of vehicles. Generally, they continuously swap the nodes order among route plans of different vehicles and also exchange the orders inside a route plan for better result. The runner-up team develops the algorithm from the perspective of threshold check, i.e., allocating the orders by checking whether they reached the threshold of delivery time and vehicle capacity, if so, those orders would be allocated. Besides, the team also implement some trick strategies like assigning the orders to the incoming vehicles for a hitch ride. The second runner-up team sets up an additional term about docking time in the optimization goal for better regulating the time cost. They also dispatch the orders based on some specific rules like urgency rank of unallocated orders.
\\
From the description above, one can notice that heuristic methods still dominate the problem among all kinds of algorithms, and the majority of them are bonded tightly to specific scenes, i.e., many tricky strategies are applied to fit for the problem settings. Although one kind of algorithm works well in one scene, it might fail to perform well when being transferred directly to another one. Therefore, generalization is definitely a challenging but valuable direction for algorithms development in the future. If such kind of algorithms was successfully found, even though it might not guarantee a global optimal solution, it would not only solve lots of realistic industrial problems, but could also help us understand the key points of the problem from theoretical perspective.
\\
We have made the tech talk videos and the source codes of awarded teams publicly available, which together with other information about this competition could be found here: \href{https://competition.huaweicloud.com/information/1000041411/introduction}{ICAPS 2021 DPDP Competition} \cite{ICAPS_DPDP2021_W3}.
\\
Though the competition is over, our work and effort on building an ecosystem based on DPDP have not finished yet. Recently, more and more research works have been focusing on machine learning aided optimization and scheduling algorithms. The combination of the advanced data-driven techniques and conventional solutions might bring a technique breakthrough in the near future for DPDP. To promote the research activities and industrial applications on this field incessantly, we decided to build the DPDP problem as a long-term competition at Huawei Competition Platform. Therefore, we also decided to keep the simulator and the dataset publicly available for any users who are interested in this field. We encourage any users, no matter you are from research institutes, universities or enterprises, to cite and use our simulator and dataset for improving the algorithm to solve DPDP problem. The long-term competition link can be found here: \href{https://competition.huaweicloud.com/information/1000041601/introduction}{Long-term DPDP Competition} \cite{Long_Term_DPDP_W3}.
\bibliographystyle{ACM-Reference-Format}
\bibliography{sample-bibliography} 


\begin{thebibliography}{10}


\ifx \showCODEN    \undefined \def \showCODEN     #1{\unskip}     \fi
\ifx \showDOI      \undefined \def \showDOI       #1{#1}\fi
\ifx \showISBNx    \undefined \def \showISBNx     #1{\unskip}     \fi
\ifx \showISBNxiii \undefined \def \showISBNxiii  #1{\unskip}     \fi
\ifx \showISSN     \undefined \def \showISSN      #1{\unskip}     \fi
\ifx \showLCCN     \undefined \def \showLCCN      #1{\unskip}     \fi
\ifx \shownote     \undefined \def \shownote      #1{#1}          \fi
\ifx \showarticletitle \undefined \def \showarticletitle #1{#1}   \fi
\ifx \showURL      \undefined \def \showURL       {\relax}        \fi
\providecommand\bibfield[2]{#2}
\providecommand\bibinfo[2]{#2}
\providecommand\natexlab[1]{#1}
\providecommand\showeprint[2][]{arXiv:#2}

\bibitem[\protect\citeauthoryear{??}{ICA}{2021}]%
        {ICAPS_DPDP2021_W3}
 \bibinfo{year}{2021}\natexlab{}.
\newblock \bibinfo{title}{ICAPS 2021 Dynamic Pickup and Delivery Problem
  Competition}.
\newblock
  \bibinfo{howpublished}{\url{https://competition.huaweicloud.com/information/1000041411/introduction}}.
    (\bibinfo{year}{2021}).
\newblock


\bibitem[\protect\citeauthoryear{??}{Lon}{2021}]%
        {Long_Term_DPDP_W3}
 \bibinfo{year}{2021}\natexlab{}.
\newblock \bibinfo{title}{Long-term Dynamic Pickup and Delivery Problem
  Competition}.
\newblock
  \bibinfo{howpublished}{\url{https://competition.huaweicloud.com/information/1000041601/introduction}}.
    (\bibinfo{year}{2021}).
\newblock


\bibitem[\protect\citeauthoryear{Gendreau, Guertin, Potvin, and
  S{\'e}guin}{Gendreau et~al\mbox{.}}{2006}]%
        {gendreau2006neighborhood}
\bibfield{author}{\bibinfo{person}{Michel Gendreau}, \bibinfo{person}{Francois
  Guertin}, \bibinfo{person}{Jean-Yves Potvin}, {and} \bibinfo{person}{Ren{\'e}
  S{\'e}guin}.} \bibinfo{year}{2006}\natexlab{}.
\newblock \showarticletitle{Neighborhood search heuristics for a dynamic
  vehicle dispatching problem with pick-ups and deliveries}.
\newblock \bibinfo{journal}{{\em Transportation Research Part C: Emerging
  Technologies\/}} \bibinfo{volume}{14}, \bibinfo{number}{3}
  (\bibinfo{year}{2006}), \bibinfo{pages}{157--174}.
\newblock


\bibitem[\protect\citeauthoryear{Ghiani, Manni, Quaranta, and Triki}{Ghiani
  et~al\mbox{.}}{2009}]%
        {ghiani2009anticipatory}
\bibfield{author}{\bibinfo{person}{Gianpaolo Ghiani}, \bibinfo{person}{Emanuele
  Manni}, \bibinfo{person}{Antonella Quaranta}, {and} \bibinfo{person}{Chefi
  Triki}.} \bibinfo{year}{2009}\natexlab{}.
\newblock \showarticletitle{Anticipatory algorithms for same-day courier
  dispatching}.
\newblock \bibinfo{journal}{{\em Transportation Research Part E: Logistics and
  Transportation Review\/}} \bibinfo{volume}{45}, \bibinfo{number}{1}
  (\bibinfo{year}{2009}), \bibinfo{pages}{96--106}.
\newblock


\bibitem[\protect\citeauthoryear{Li, Luo, Yuan, Wang, Lu, Wang, Lu, and
  Zeng}{Li et~al\mbox{.}}{2021}]%
        {li2021learning}
\bibfield{author}{\bibinfo{person}{Xijun Li}, \bibinfo{person}{Weilin Luo},
  \bibinfo{person}{Mingxuan Yuan}, \bibinfo{person}{Jun Wang},
  \bibinfo{person}{Jiawen Lu}, \bibinfo{person}{Jie Wang},
  \bibinfo{person}{Jinhu Lu}, {and} \bibinfo{person}{Jia Zeng}.}
  \bibinfo{year}{2021}\natexlab{}.
\newblock \bibinfo{title}{Learning to Optimize Industry-Scale Dynamic Pickup
  and Delivery Problems}.
\newblock   (\bibinfo{year}{2021}).
\newblock
\showeprint[arxiv]{cs.AI/2105.12899}


\bibitem[\protect\citeauthoryear{Ma, Hao, HAO, Lu, Liu, Tong, Yuan, Li, Tang,
  and Meng}{Ma et~al\mbox{.}}{2021}]%
        {ma2021a}
\bibfield{author}{\bibinfo{person}{Yi Ma}, \bibinfo{person}{Xiaotian Hao},
  \bibinfo{person}{Jianye HAO}, \bibinfo{person}{Jiawen Lu},
  \bibinfo{person}{Xing Liu}, \bibinfo{person}{Xialiang Tong},
  \bibinfo{person}{Mingxuan Yuan}, \bibinfo{person}{Zhigang Li},
  \bibinfo{person}{Jie Tang}, {and} \bibinfo{person}{Zhaopeng Meng}.}
  \bibinfo{year}{2021}\natexlab{}.
\newblock \showarticletitle{A Hierarchical Reinforcement Learning Based
  Optimization Framework for Large-scale Dynamic Pickup and Delivery Problems}.
  \bibinfo{howpublished}{\url{https://openreview.net/forum?id=2F_wnaioS6}}. In
  \bibinfo{booktitle}{{\em Thirty-Fifth Conference on Neural Information
  Processing Systems}}.
\newblock


\bibitem[\protect\citeauthoryear{Mitrovic-Minic, Krishnamurti, Laporte,
  et~al\mbox{.}}{Mitrovic-Minic et~al\mbox{.}}{2004}]%
        {mitrovic2004double}
\bibfield{author}{\bibinfo{person}{Snezana Mitrovic-Minic},
  \bibinfo{person}{Ramesh Krishnamurti}, \bibinfo{person}{Gilbert Laporte},
  {et~al\mbox{.}}} \bibinfo{year}{2004}\natexlab{}.
\newblock \showarticletitle{Double-horizon based heuristics for the dynamic
  pickup and delivery problem with time windows}.
\newblock \bibinfo{journal}{{\em Transportation Research Part B:
  Methodological\/}} \bibinfo{volume}{38}, \bibinfo{number}{8}
  (\bibinfo{year}{2004}), \bibinfo{pages}{669--685}.
\newblock


\bibitem[\protect\citeauthoryear{Savelsbergh and Sol}{Savelsbergh and
  Sol}{1998}]%
        {savelsbergh1998drive}
\bibfield{author}{\bibinfo{person}{Martin Savelsbergh} {and}
  \bibinfo{person}{Marc Sol}.} \bibinfo{year}{1998}\natexlab{}.
\newblock \showarticletitle{Drive: Dynamic routing of independent vehicles}.
\newblock \bibinfo{journal}{{\em Operations Research\/}} \bibinfo{volume}{46},
  \bibinfo{number}{4} (\bibinfo{year}{1998}), \bibinfo{pages}{474--490}.
\newblock


\bibitem[\protect\citeauthoryear{Thomas}{Thomas}{2007}]%
        {thomas2007waiting}
\bibfield{author}{\bibinfo{person}{Barrett~W Thomas}.}
  \bibinfo{year}{2007}\natexlab{}.
\newblock \showarticletitle{Waiting strategies for anticipating service
  requests from known customer locations}.
\newblock \bibinfo{journal}{{\em Transportation Science\/}}
  \bibinfo{volume}{41}, \bibinfo{number}{3} (\bibinfo{year}{2007}),
  \bibinfo{pages}{319--331}.
\newblock


\bibitem[\protect\citeauthoryear{Toth and Vigo}{Toth and Vigo}{2014}]%
        {toth2014vehicle}
\bibfield{author}{\bibinfo{person}{Paolo Toth} {and} \bibinfo{person}{Daniele
  Vigo}.} \bibinfo{year}{2014}\natexlab{}.
\newblock \bibinfo{booktitle}{{\em Vehicle routing: problems, methods, and
  applications}}.
\newblock \bibinfo{publisher}{SIAM}.
\newblock


\end{thebibliography}

\end{sloppypar}
\end{document}